%% file: main.tex
\documentclass[10pt,twocolumn,letterpaper]{article}

\PassOptionsToPackage{table}{xcolor}  
\usepackage[pagenumbers]{iccv} 
\usepackage{amsmath}
\usepackage{amssymb}
\usepackage{natbib}
\usepackage{graphicx}
\usepackage{times}
\usepackage{epsfig}
\usepackage{algorithm}
\usepackage{algpseudocode}
\usepackage{booktabs}
\usepackage{tabularx}

\newif\ifreview
\reviewfalse
\makeatletter
\@ifpackagewith{iccv}{review}{
    \reviewtrue
}{}
\makeatother

\input{preamble}

\definecolor{iccvblue}{rgb}{0.21,0.49,0.74}
\usepackage[pagebackref,breaklinks,colorlinks,allcolors=iccvblue]{hyperref}

\def\paperID{14936} 
\def\confName{ICCV}
\def\confYear{2025}

\title{Speedy MASt3R}

\author{Jingxing Li\thanks{Equal contribution.}\\
Arizona State University\\
Tempe, AZ, USA\\
{\tt\small jingxing@asu.edu}
\and
Yongjae Lee\footnotemark[1]\\
Arizona State University\\
Tempe, AZ, USA\\
{\tt\small ylee298@asu.edu}
\and
Abhay Kumar Yadav\\
Johns Hopkins University\\
Baltimore, MD, USA\\
{\tt\small ayadav13@jh.edu}
\and
Cheng Peng\\
Johns Hopkins University\\
Baltimore, MD, USA\\
{\tt\small cpeng26@jhu.edu}
\and
Rama Chellappa\\
Johns Hopkins University\\
Baltimore, MD, USA\\
{\tt\small rchella4@jhu.edu}
\and
Deliang Fan\footnotemark[2]\\
Arizona State University\\
Tempe, AZ, USA\\
{\tt\small dfan@asu.edu}
}

\begin{document}
\maketitle

\begin{abstract}
Image matching is a fundamental component of state-of-the-art 3D vision algorithms and pipelines, playing a crucial role in accurate scene reconstruction and localization. MASt3R~\cite{leroy2024groundingimagematching3d}  has redefined image matching as a 3D task by leveraging DUSt3R~\cite{wang2024dust3rgeometric3dvision} and introducing a fast reciprocal matching scheme that accelerates matching by orders of magnitude while maintaining theoretical guarantees. This approach has gained significant traction in the community, with DUSt3R and MASt3R collectively accumulating over 250 citations in a short span, underscoring their impact. However, despite its state-of-the-art accuracy, MASt3R's inference speed remains a bottleneck, for example on an A40 GPU, with a latency of 198.16 ms per image pair, primarily due to computational overhead from the ViT encoder-decoder and the Fast Reciprocal Nearest Neighbor (FastNN) matching stage.

To address this, we introduce \textbf{Speedy MASt3R}, a post-training optimization framework that significantly enhances inference efficiency while maintaining accuracy. Speedy MASt3R integrates multiple optimization techniques, including FlashMatch---an approach that leverages FlashAttention~v2 with tiling strategies to significantly enhance computational efficiency---computation graph optimization with layer and tensor fusion, kernel auto-tuning via TensorRT (GraphFusion), and a streamlined FastNN pipeline that reduces memory access time from quadratic to linear while accelerating block-wise correlation scoring through vectorized computation~(FastNN-Lite). Additionally, it employs mixed-precision inference with FP16/FP32 hybrid computations (HybridCast), achieving speedup while ensuring numerical precision. Evaluated on Aachen Day-Night, InLoc, 7-Scenes, ScanNet1500, and MegaDepth1500 datasets, Speedy MASt3R achieves a \textbf{54\% reduction in inference time} (198 ms $\to$ 91 ms per image pair) without compromising accuracy. This advancement enables real-time 3D understanding, facilitating applications such as mixed reality navigation and large-scale 3D scene reconstruction.
\end{abstract}

\section{Introduction}
\label{sec:intro}

Image matching is a fundamental problem in computer vision, crucial for applications such as structure-from-motion (SfM)~\cite{schonberger2016structure}, visual localization~\cite{taira2018inloc, sattler2018benchmarking}, and 3D reconstruction~\cite{agarwal2011building, fuhrmann2014mve}. Traditional keypoint-based methods, including SIFT~\cite{lowe2004distinctive}, ORB~\cite{rublee2011orb}, and SuperPoint~\cite{detone2018superpoint}, detect and describe sparse features before performing nearest-neighbor search for matching. While these methods remain effective in many scenarios, their reliance on local descriptors makes them vulnerable to texture-less regions and repetitive patterns.

To overcome these limitations, deep learning-based dense matching techniques, such as LoFTR~\cite{sun2021loftr}, DKM~\cite{edstedt2023dkm}, RoMa~\cite{Edstedt2024RoMa}, and SuperGlue~\cite{sarlin2020superglue}, leverage global feature reasoning through transformer-based architectures. These methods achieve state-of-the-art performance on challenging benchmarks, improving robustness to large viewpoint and illumination changes. However, dense matching often incurs high computational costs, making it less feasible for real-time applications.

More recently, grounding image matching in 3D has gained attention as a means to improve both robustness and accuracy. DUSt3R~\cite{wang2024dust3rgeometric3dvision} pioneered the use of 3D pointmaps for pixel correspondences, demonstrating superior resilience to extreme viewpoint variations. MASt3R~\cite{leroy2024groundingimagematching3d} extends this approach by integrating a transformer-based matching head that learns local features alongside the 3D structure, enabling more precise matches. Our work, Speedy MASt3R builds upon this foundation, introducing computational-efficiency attention mechanisms~\cite{dao2023flashattention2fasterattentionbetter} and computational graph optimizations~\cite{tensorrt} to accelerate inference while maintaining accuracy. Our approach preserves the theoretical guarantees of the fast reciprocal matching scheme used in the original MASt3R while reducing memory access times and enhancing computational efficiency, enabling real-time performance without sacrificing accuracy.
Our work, {Speedy MASt3R}, introduces a comprehensive post-training optimization framework to accelerate image matching while maintaining state-of-the-art accuracy. It integrates several major optimization techniques:
\begin{itemize}
    \item \textbf{FlashMatch}: An efficient attention mechanism leveraging {FlashAttention v2}~\cite{dao2023flashattention2fasterattentionbetter} with tiling strategies to optimize GPU memory access and significantly reduce computational overhead in the ViT encoder-decoder pipeline~\cite{Dosovitskiy2021Image}.
    \item \textbf{GraphFusion}: Computation graph optimization by utilizing kernel auto-tuning and tensor fusion, eliminating redundant intermediate tensor allocations and reducing unnecessary computations, as leveraged by TensorRT~\cite{tensorrt}.
    \item \textbf{FastNN-Lite}: A streamlined FastNN pipeline that reduces memory access time from quadratic to linear and accelerates block-wise correlation scoring through vectorized computation.
    \item \textbf{HybridCast}: A mixed-precision inference framework combining {FP16} and {FP32} computations to achieve speedup while ensuring numerical precision in critical operations.
\end{itemize}

Speedy MASt3R achieves a \textbf{54\% reduction in inference time} (198 ms → 91 ms per image pair) without compromising high quality matching results, as demonstrated on the {Aachen Day-Night}~\cite{Zhang2021Reference}, {InLoc}~\cite{taira2018inloc}, {7-Scenes}~\cite{shotton2013scene}, {ScanNet1500}~\cite{dai2017scannet} and {MegaDepth1500}~\cite{li2018megadepth} datasets datasets. This significant speedup underscores the effectiveness of our optimization framework in enabling real-time 3D understanding without sacrificing performance.

\section{Background and Related Works}
\label{sec:related_work}

Recent advancements in image matching have redefined the landscape of 3D scene reconstruction and visual localization. Traditional methods such as SIFT~\cite{lowe2004distinctive} and ORB~\cite{rublee2011orb} rely on handcrafted keypoints and descriptors, making them susceptible to texture-less surfaces and extreme viewpoint changes. Learning-based methods such as SuperPoint~\cite{detone2018superpoint} and SuperGlue~\cite{sarlin2020superglue} improve feature matching by leveraging deep neural networks and global feature aggregation. However, they still treat matching as a local problem, which can lead to inconsistencies in large-scale 3D scene reconstruction.

\subsection{MASt3R and 3D-Grounded Matching}
To address these challenges, DUSt3R~\cite{wang2024dust3rgeometric3dvision} introduced 3D pointmaps, which frame image matching as a joint 3D scene reconstruction problem. Extending this idea, MASt3R~\cite{leroy2024groundingimagematching3d} introduced a transformer-based matching head that jointly learns local features and 3D correspondences. Additionally, Fast Nearest-Neighbor Matching (FastNN) was proposed as a high-efficiency nearest-neighbor search mechanism. MASt3R achieved state-of-the-art performance on multiple benchmarks, demonstrating robustness to extreme viewpoint changes. Despite these innovations, MASt3R's inference speed remains a bottleneck, primarily due to its heavy computation from the ViT encoder-decoder, which accounts for 60\% of the latency, and the FastNN matching stage, which contributes to 40\% of total computation time. Moreover, the significant computational overhead associated with full-resolution dense correspondences renders it impractical for real-time applications, such as AR/VR, robotics, and large-scale mapping. Resolving these computational bottlenecks is essential for enabling practical deployment in time-sensitive scenarios.

\subsection{Optimizing Image Matching for Speed and Efficiency}
Several recent works have focused on optimizing dense feature matching for efficiency. Vision transformers (ViTs)~\cite{Dosovitskiy2021Image} have been a critical development in global feature aggregation. Swin Transformer~\cite{liu2021swin} reduces computational complexity by restricting self-attention to local windows, making transformers more scalable for high-resolution images. FlashAttention~\cite{dao2022flashattention} and FlashAttention v2~\cite{dao2023flashattention2fasterattentionbetter} further optimize GPU memory access by introducing tiling strategies. These improvements allow for efficient sequence processing without compromising accuracy.

\subsection{Efficient Attention Mechanisms and FlashAttention}
Traditional self-attention mechanisms in transformers suffer from quadratic complexity with respect to sequence length, making them inefficient for large-scale feature matching tasks. FlashAttention~\cite{dao2022flashattention} optimizes memory access by using an I/O-aware algorithm that avoids materializing the full attention matrix, significantly reducing both computation and memory costs. It achieves this by tiling the attention computation, ensuring that intermediate values fit within high-bandwidth memory (SRAM) on GPUs. FlashAttention v2~\cite{dao2023flashattention2fasterattentionbetter} improves upon this by further optimizing work partitioning and parallelization, achieving significant speedup compared to naive attention implementations. 

\subsection{Efficient Nearest-Neighbor Search for Feature Matching}
Efficient nearest-neighbor search remains a key challenge in large-scale feature matching. Traditional mutual nearest neighbor search methods have quadratic complexity, making them infeasible for dense matching. Faiss~\cite{johnson2019billion} addresses this by employing approximate nearest-neighbor (ANN) search, enabling large-scale similarity retrieval. Similarly, HNSW graphs~\cite{malkov2020hnsw} optimize nearest-neighbor retrieval using multi-layer navigable small-world graphs, but these methods are not accurate. FastNN, introduced in MASt3R, aimed to reduce this computational overhead while preserving accuracy, but it still remains a bottleneck in dense matching pipelines.

\subsection{Mixed-Precision and Kernel Fusion for Speedup}
Further acceleration can be achieved through mixed-precision inference and computational graph optimizations. TensorRT~\cite{tensorrt}-based optimizations eliminate redundant intermediate tensor allocations, thereby reducing unnecessary computations. Additionally, mixed-precision inference (FP16/FP32) has been shown to significantly reduce memory bandwidth. Such optimizations allow models to achieve substantial speedups while preserving performance for critical tasks like 3D scene reconstruction and image matching.

\begin{figure*}[t]
\centering
\includegraphics[width=\textwidth]{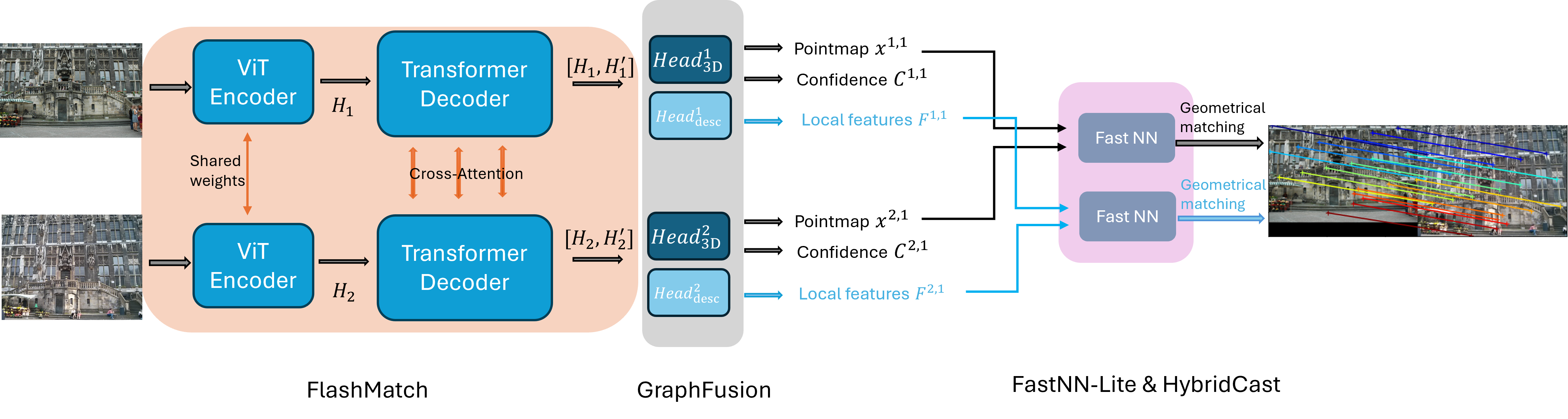}
\caption{Overview of the MASt3R pipeline and optimizations introduced by Speedy MASt3R. Given two input images, the network leverages a ViT encoder and a transformer decoder to jointly regress 3D pointmaps, confidence maps, and dense feature maps. The FastNN matcher identifies robust correspondences, enabling joint 3D reconstruction and image matching. Speedy MASt3R enhances the original framework by integrating FlashMatch for efficient attention computation through tiling strategies, GraphFusion for eliminating redundant, unnecessary tensor computation, FastNN-Lite for reducing memory access time from quadratic to linear, and HybridCast for enabling mixed-precision inference with FP16 and FP32 computations.}
\label{fig:mast3r}
\end{figure*}

\section{Method}
\label{sec:method}

\subsection{Problem Statement}
Given two images \(I_1\) and \(I_2\), captured by two cameras \(c_1\) and \(c_2\) with unknown parameters, the goal is to recover a set of pixel correspondences \(\{(i, j)\}\), where \(i\) and \(j\) are pixels in \(I_1\) and \(I_2\) respectively. Each pixel is represented as \(i = (w_i, h_i)\) and \(j = (w_j, h_j)\), where \(w\) and \(h\) denote the width and height of the images. For simplicity, \(I_1\) and \(I_2\) are assumed to have the same resolution, although the approach can handle pairs of variable aspect ratios.

The problem of image matching is inherently tied to the recovery of 3D scene geometry. Traditional methods cast matching as a 2D problem, which limits their applicability for tasks like visual localization. In contrast, MASt3R \cite{leroy2024groundingimagematching3d} jointly addresses 3D scene reconstruction and image matching, leveraging the DUSt3R\cite{wang2024dust3rgeometric3dvision} framework  as a foundation.

\subsection{Overview of MASt3R}
MASt3R, illustrated in Figure~\ref{fig:mast3r}, builds upon the DUSt3R framework and introduces a novel matching head and an optimized matching scheme. The pipeline consists of the following key steps:

\subsubsection{Feature Extraction}
Both images \(I_1\) and \(I_2\) are encoded in a Siamese manner using CroCo \cite{weinzaepfel2023crocoselfsupervisedpretraining3d}, which is a Vision Transformer (ViT), yielding two representations \(H_1\) and \(H_2\):
\[
H_1, H_2 = \text{Encoder}(I_1), \text{Encoder}(I_2).
\]

\subsubsection{Cross-Attention Decoding}
The representations \(H_1\) and \(H_2\) are processed by two intertwined decoders, which also utilize the CroCo \cite{weinzaepfel2023crocoselfsupervisedpretraining3d} structure. These decoders exchange information via cross-attention to understand the spatial relationship between viewpoints and the global 3D geometry of the scene. The augmented representations are denoted as \(H_1'\) and \(H_2'\):
\[
H_1', H_2' = \text{Decoder}(H_1, H_2).
\]

\subsubsection{3D Pointmap Regression}
Two prediction heads regress dense 3D pointmaps \(x_{1,1}\) and \(x_{2,1}\), as well as confidence maps \(c_1\) and \(c_2\):
\[
x_{1,1}, c_1 = \text{Head}_{p}([H_1, H_1']),
\]
\[
x_{2,1}, c_2 = \text{Head}_{p}([H_2, H_2']).
\]
Here, \([H_1, H_1']\) and \([H_2, H_2']\) are the concatenations of the encoder and decoder outputs. \(x_{1,1} \in \mathbb{R}^{H \times W \times 3}\) represents a dense 2D-to-3D mapping between each pixel in \(I_1\) and its corresponding 3D point in the coordinate system of camera \(c_1\).

\subsubsection{Matching Head}
To improve the precision of pixel correspondences, MASt3R introduces a matching head that outputs dense feature maps \(D_1\) and \(D_2 \in \mathbb{R}^{H \times W \times d}\):
\[
D_1, D_2 = \text{Head}_{m}([H_1, H_1']), \text{Head}_{m}([H_2, H_2']).
\]
These feature maps are used in conjunction with the 3D pointmaps to perform robust matching.

\subsubsection{Fast Reciprocal NN Matching}
MASt3R introduces an optimized matching scheme based on \textbf{Fast Reciprocal NN Matching (FastNN)} to efficiently handle dense feature maps. This scheme is designed to reduce computational complexity while maintaining high matching accuracy, making it suitable for large-scale datasets.

\paragraph{Problem Context}
Traditional mutual nearest neighbor (NN) matching methods require computing pairwise distances between all pixels, resulting in a complexity of \[O(W^2H^2)\], where \(W\) and \(H\) are the width and height of the images. This high complexity becomes a bottleneck for large-scale datasets and real-time applications.

\paragraph{FastNN Algorithm}
FastNN addresses this issue by leveraging iterative subsampling and reciprocal NN search. The algorithm proceeds as follows:
\begin{enumerate}
    \item \textbf{Initialization}: Sample \(k\) pixels \(U^0\) from \(I_1\) typically on a grid. Find their nearest neighbors in \(I_2\), denoted as \(V^0\).
    \item \textbf{Iterative Search}: In each iteration \(t\), find the nearest neighbors of \(V^t\) back in \(I_1\), denoted as \(U^{t+1}\). Identify reciprocal matches \(M_t = \{(i, j) \mid U^{t+1}_i = U^t_i\}\) (points forming a cycle). Remove converged points from \(U^{t+1}\) and \(V^{t+1}\).
    \item \textbf{Termination}: The process terminates when most points have converged or a maximum number of iterations \(T\) is reached.
    \item \textbf{Output}: Return the set of all reciprocal matches \(M = \bigcup_t M_t\).
\end{enumerate}

\paragraph{Integration with MASt3R}
In MASt3R, FastNN is applied in a coarse-to-fine manner to improve both speed and accuracy. The dense feature maps \(D_1\) and \(D_2\) generated by the matching head are used as input to FastNN. This allows MASt3R to efficiently compute robust pixel correspondences, which are then used for 3D reconstruction.

\subsubsection{3D Reconstruction}
Finally, the dense correspondences are used to generate a 3D point cloud, leveraging the DUSt3R framework's regression loss for optimization.

\subsection{Limitations of MASt3R}
While MASt3R achieves state-of-the-art accuracy in 3D scene reconstruction and image matching, its inference speed remains a bottleneck. Specifically, processing a single image pair takes 198ms, which is significantly slower than real-time requirements. This slow matching speed severely limits the real-time applicability of MASt3R, particularly in scenarios requiring fast and efficient processing, such as autonomous driving or augmented reality.

To address these challenges, Speedy MASt3R is proposed as an optimized framework that significantly reduces inference latency without compromising accuracy. The following sections detail the key optimizations introduced in Speedy MASt3R to overcome the limitations of the original MASt3R pipeline.

\subsection{Speedy MASt3R}
\subsubsection{FlashMatch}
\label{sec:memory_efficient_attention}
The Vision Transformer (ViT)~\cite{Dosovitskiy2021Image} encoder-decoder in MASt3R plays a crucial role in 3D scene reconstruction and image matching. However, the traditional attention mechanism in ViT suffers from high computational complexity, scaling quadratically with the sequence length of the input tokens \(O(n^2)\), and a significant memory footprint. This becomes a bottleneck for MASt3R, as 60\% of the total inference latency is attributed to the ViT encoder-decoder, with attention being the primary contributor. Specifically, the memory-intensive nature of attention computation limits the scalability of MASt3R to high-resolution images and real-time applications.

To address these limitations, we integrate {FlashAttention v2} \cite{dao2023flashattention2fasterattentionbetter} into the self-attention modules of 2 pairs of encoders and decoders in MASt3R. FlashAttention v2 is an optimized attention mechanism that reduces both computational complexity and memory footprint by leveraging tiling strategies and efficient memory access patterns. Its core idea is to decompose the attention computation into smaller blocks (tiles) that fit into the GPU's fast memory (SRAM), minimizing the need for costly global memory accesses.

\begin{figure*}[t]
\centering
\includegraphics[width=.9\textwidth]{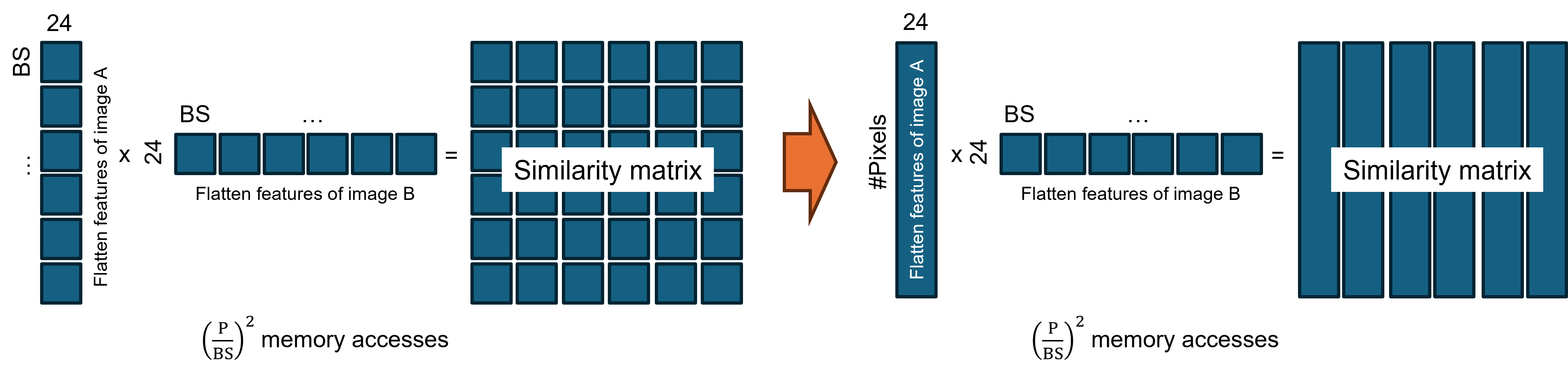}
\caption{Comparison of Double Loop (left) and Single Loop (right) optimization strategies for matrix multiplication in the feature matching stage of MASt3R. Here, BS denotes Block Size, and P denotes the number of Pixels. The traditional Double Loop approach incurs significant memory access overhead due to block-wise computation. Our proposed Single Loop strategy unrolls block-wise operations into a single loop, reducing memory accesses while maintaining VRAM usage within the target hardware's capacity.}
\label{fig:overview}
\end{figure*}

\subsubsection{GraphFusion}
While FlashMatch enhances inference speed by optimizing the Attention mechanism in the Transformer, we further accelerate the entire network's execution by applying several inference-time optimization techniques. These include computation graph optimization, layer and tensor fusion, efficient memory management for dynamic tensors, and kernel tuning for deployment target device. Leveraging the TensorRT~\cite{tensorrt}, we achieve more efficient computational graph fusion and optimization, significantly boosting inference speed of neural network.

\subsubsection{FastNN-Lite}
The original FastNN employs a nested loop structure to compute pairwise distances between feature blocks from two images \(A\) and \(B\), resulting in \(O(n^2)\) time complexity. The algorithm is formalized as follows:

\begin{algorithm}[t]
\small 
\caption{Original FastNN}
\begin{algorithmic}[1]
\State \textbf{Input:} Feature blocks \(A\) and \(B\)
\State \textbf{Output:} Nearest neighbors for each block in \(A\)
\State \(N_A \gets \text{len}(A)\) \Comment{Number of blocks in \(A\)}
\State \(N_B \gets \text{len}(B)\) \Comment{Number of blocks in \(B\)}
\State \texttt{nearest\_neighbors} \(\gets []\) \Comment{Store nearest neighbors}
\For{$i = 1$ to \(N_A$} \Comment{Outer loop over \(A\)}
    \State \(A_{\text{block}} \gets A[i]\) \Comment{Extract current block from \(A\)}
    \State \(\text{min\_dist} \gets \infty\) \Comment{Initialize minimum distance}
    \State \(\text{nearest\_idx} \gets -1\) \Comment{Initialize nearest neighbor index}
    \For{$j = 1$ to \(N_B$} \Comment{Inner loop over \(B\)}
        \State \(B_{\text{block}} \gets B[j]\) \Comment{Extract current block from \(B\)}
        \State \(\text{dist} \gets \text{dist\_func}(A_{\text{block}}, B_{\text{block}})\) \Comment{Compute pairwise distance}
        \If{\(\text{dist} < \text{min\_dist}\)} \Comment{Update nearest neighbor}
            \State \(\text{min\_dist} \gets \text{dist}\)
            \State \(\text{nearest\_idx} \gets j\)
        \EndIf
    \EndFor
    \State \(\texttt{nearest\_neighbors.append}(\text{nearest\_idx})\) \Comment{Store result}
\EndFor
\State \textbf{return} \texttt{nearest\_neighbors}
\end{algorithmic}
\end{algorithm}

We notice that the original FastNN algorithm can be sped up by reducing the number of accesses to the feature blocks. Therefore, we suggest substituting the original algorithm with FastNN-Lite. FastNN-Lite first replaces the nested loop structure with a single-loop execution graph, processing blocks of \(A\) sequentially while handling \(B\) as a whole. This modification reduces the time complexity to \(O(n)\) and eliminates redundant memory allocations. The algorithm is formalized as follows:

\begin{algorithm}[t]
\small 
\caption{FastNN-Lite}
\begin{algorithmic}[1]
\State \textbf{Input:} Feature blocks \(A\) and \(B\)
\State \textbf{Output:} Nearest neighbors for each block in \(A\)
\State \(N_A \gets \text{len}(A)\) \Comment{Number of blocks in \(A\)}
\State \texttt{nearest\_neighbors} \(\gets []\) \Comment{Store nearest neighbors}
\For{$i = 1$ to \(N_A$} \Comment{Single loop over \(A\)}
    \State \(A_{\text{block}} \gets A[\text{start}_i : \text{end}_i]\) \Comment{Extract current block from \(A\)}
    \State \(\text{dists}_{\text{blk}} \gets \text{dist\_func}(A_{\text{block}}, B)\) \Comment{Compute distances to all blocks in \(B\)}
    \State \(\text{nearest\_idx} \gets \text{argmin}(\text{dists}_{\text{blk}})\) \Comment{Find nearest neighbor}
    \State \(\texttt{nearest\_neighbors.append}(\text{nearest\_idx})\) \Comment{Store result}
\EndFor
\State \textbf{return} \texttt{nearest\_neighbors}
\end{algorithmic}
\end{algorithm}

\paragraph{Key Optimizations.}
The single-loop execution graph introduces several significant optimizations, as illustrated in Figure~\ref{fig:overview}:
\begin{itemize}
    \item \textbf{Time Complexity Reduction}: By processing blocks of \(A\) sequentially and handling \(B\) as a whole, the time complexity is reduced from \(O(n^2)\) to \(O(n)\).
    \item \textbf{Efficient Memory Allocation}: The FastNN-Lite approach eliminates redundant memory allocations by avoiding intermediate storage for pairwise comparisons. This is achieved through the single-loop optimization strategy depicted in Figure~\ref{fig:overview}, which reduces memory accesses from \((\text{P}/\text{BS})^2\) to \((\text{P}/\text{BS})\). Here, BS denotes Block Size, and P denotes the number of Pixels.
    \item \textbf{Vectorized Computation}: The distance function \(\text{dist\_func}\) is applied in a vectorized manner, enabling efficient computation of distances between \(A_{\text{block}}\) and all blocks in \(B\) simultaneously. This aligns with the single-loop approach shown in Figure~\ref{fig:overview}, further enhancing computational efficiency.
\end{itemize}

\subsubsection{HybridCast}
HybridCast leverages both \textbf{FP16} (16-bit floating point) and \textbf{FP32} (32-bit floating point) to optimize computational efficiency and memory usage without sacrificing model accuracy. This technique is particularly effective in deep learning tasks where memory bandwidth and computational speed are critical bottlenecks.

In our implementation, HybridCast is applied during the \textbf{feature matching stage} of the MASt3R pipeline. Specifically, we use FP16 for \textbf{distance computation} and FP32 for \textbf{gradient accumulation} and \textbf{final result aggregation}. This approach reduces memory footprint and accelerates computation while maintaining numerical stability.

The formalized algorithm for HybridCast in the context of feature matching is as follows:

\begin{algorithm}[t]
\small 
\caption{HybridCast}
\begin{algorithmic}[1]
\State \textbf{Input:} Feature blocks \(A\) and \(B\), distance function \(\text{dist\_func}\)
\State \textbf{Output:} Nearest neighbors for each block in \(A\)
\State \(N_A \gets \text{len}(A)\) \Comment{Number of blocks in \(A\)}
\State \texttt{nearest\_neighbors} \(\gets []\) \Comment{Store nearest neighbors}

\For{$i = 1$ to \(N_A$} \Comment{Loop over blocks in \(A\)}
    \State \(A_{\text{block}} \gets A[\text{start}_i : \text{end}_i]\) \Comment{Extract current block from \(A\)}
    \State \(A_{\text{block}}^{\text{FP16}} \gets \text{FP16}(A_{\text{block}})\) \Comment{Convert \(A_{\text{block}}\) to FP16}
    \State \(B^{\text{FP16}} \gets \text{FP16}(B)\) \Comment{Convert \(B\) to FP16}
    \State \(\text{dists}_{\text{blk}} \gets \text{dist\_func}(A_{\text{block}}^{\text{FP16}}, B^{\text{FP16}})\) \Comment{Compute distances in FP16}
    \State \(\text{dists}_{\text{blk}}^{\text{FP32}} \gets \text{FP32}(\text{dists}_{\text{blk}})\) \Comment{Convert distances to FP32 for aggregation}
    \State \(\text{nearest\_idx} \gets \text{argmin}(\text{dists}_{\text{blk}}^{\text{FP32}})\) \Comment{Find NN in FP32}
    \State \(\texttt{nearest\_neighbors.append}(\text{nearest\_idx})\) \Comment{Store result}
\EndFor

\State \textbf{return} \texttt{nearest\_neighbors}
\end{algorithmic}
\end{algorithm}

HybridCast is applied in the MASt3R pipeline to optimize performance and resource utilization. During \textbf{feature extraction}, feature maps are converted to FP16, reducing memory usage by 50\% and enabling larger batch sizes. In the \textbf{distance computation} stage, pairwise distance calculations are performed in FP16, leveraging the accelerated computation capabilities of modern GPUs to achieve up to 2x speedup. Finally, for \textbf{result aggregation}, distances are converted back to FP32 to ensure numerical stability and accurate nearest neighbor selection, maintaining negligible accuracy loss. This approach combines the efficiency of FP16 with the precision of FP32, delivering significant performance improvements while minimizing resource overhead. Notably, if one naively uses only FP16, it can lead to poor performance, which is difficult and non-trivial to diagnose due to subtle numerical instabilities affecting nearest-neighbor selection and gradient computations.

\section{Experiments}
\label{sec:experiments}
Speedy MASt3R is a post-training optimization framework. We base the MASt3R's architecture (ViT-Large encoder, ViT-Base decoder, and CatMLP+DPT head) and initialize with the public pretrained weights. Then, we directly apply the optimization techniques.

We evaluate our proposed \textbf{Speedy MASt3R} on the two popular tasks with widely used benchmarks. For the relative pose estimation task~(\cref{sec:rel_pose_est}), we report results on the ScanNet1500~\cite{sarlin2020superglue,dai2017scannet} and MegaDepth1500~\cite{li2018megadepth,sun2021loftr} datasets. For the visual localization task~(\cref{sec:visual_localization}), we present results on the Aachen Day-Night~\cite{Zhang2021Reference}, InLoc~\cite{taira2018inloc}, and 7-Scenes~\cite{shotton2013scene} datasets. We conducted our experiment on an A40 GPU.

\subsection{Relative Pose Estimation}
\label{sec:rel_pose_est}
We evaluate Speedy MASt3R on the ScanNet1500~\cite{dai2017scannet} and MegaDepth1500~\cite{li2018megadepth} datasets. Both datasets contain 1,500 pairs of images, with ScanNet1500 focusing more on indoor images, while MegaDepth1500 consists exclusively of outdoor images. We report model accuracy using four metrics: AUC@5/10/20, which measures the area under the curve of pose accuracy with respect to thresholds of 5/10/20 degrees for the minimum of translation and rotation angular errors, and mean average accuracy (mAA), which is the mean of AUC@5/10/20. Additionally, we measure the average running time of each module (Encoder/Decoder/Head/FastNN) in milliseconds (ms).


\Cref{tab:1} compares Speedy MASt3R with vanilla MASt3R in terms of accuracy and computational efficiency. While maintaining the same accuracy—since the difference is not statistically significant—the optimization techniques effectively reduce the running time of each module by 47.41\%, 30.99\%, 26.73\%, 61.07\% for ScanNet1500 and by 47.12\%, 30.41\%, 27.11\%, 58.96\% for MegaDepth1500. 

\begin{table*}[]
\centering
\caption{Accuracy (left) and computational efficiency (right) test on ScanNet1500~\cite{dai2017scannet} and MegaDepth1500~\cite{li2018megadepth} datasets. Lower numbers are better in terms of inference speed.}
\label{tab:1}
\begin{minipage}{0.6\textwidth}
\centering
\resizebox{\textwidth}{!}{%
\begin{tabular}{@{}lcccccccc@{}}
\toprule
\multirow{2}{*}{}          & \multicolumn{4}{c}{ScanNet1500~\cite{dai2017scannet}} & \multicolumn{4}{c}{MegaDepth1500~\cite{li2018megadepth}} \\ \cmidrule(l){2-9} 
                           & AUC@5 & AUC@10 & AUC@20 & mAA   & AUC@5  & AUC@10  & AUC@20 & mAA   \\ \midrule
\multicolumn{1}{c}{MASt3R} & 34.51 & 57.31  & 74.5   & 55.44 & 39.87  & 54.93   & 66.88  & 53.89 \\
\multicolumn{1}{c}{Ours} & 34.32 & 57.15  & 74.3   & 55.25 & 39.28  & 54.57   & 66.54  & 53.46 \\ \bottomrule
\end{tabular}%
}
\end{minipage}
\hfill
\begin{minipage}{0.39\textwidth}

\centering
\resizebox{\textwidth}{!}{%
\begin{tabular}{@{}lcccccc@{}}
\toprule
                              & \multicolumn{1}{l}{} & Encoder\(\downarrow\) & Decoder\(\downarrow\) & Head\(\downarrow\) & FastNN\(\downarrow\)       \\ \midrule
\multicolumn{1}{c}{ScanNet}   & MASt3R               & 52.79  & 31.01  & 20.72            & 115.69 \\
                              & Ours                 &  \textbf{27.76}  & \textbf{ 21.40}  &  \textbf{15.18}            &  \textbf{45.03}   \\ \midrule
\multicolumn{1}{c}{MegaDepth} & MASt3R               & 52.34  & 30.85  & 20.77            & 114.17 \\
                              & Ours                 &  \textbf{27.68}  &  \textbf{21.47}  &  \textbf{15.14}            &  \textbf{46.86}    \\ \bottomrule
\end{tabular}%
}
\end{minipage}

\end{table*}

\begin{table*}[]
\centering
\caption{Localization Accuracies. The upper table presents the percentage of accurately localized images within the thresholds of (0.25m/2\textdegree)/(0.5m/5\textdegree)/(5m/10\textdegree) for Aachen~\cite{Zhang2021Reference}, and (0.25m/10\textdegree)/(0.5m/10\textdegree)/(1m/10\textdegree) for InLoc~\cite{taira2018inloc}. The lower table reports localization accuracy using median translation and rotation errors for 7-Scenes~\cite{shotton2013scene}. The ``top N'' indicates the number of retrieved images.}
\label{tab:2}
\resizebox{\textwidth}{!}{%
\begin{tabular}{@{}ccccc@{}}
\toprule
\multicolumn{1}{l}{\multirow{2}{*}{}} & \multicolumn{2}{c}{Aachen~\cite{Zhang2021Reference}} & \multicolumn{2}{c}{InLoc~\cite{taira2018inloc}} \\ \cmidrule(l){2-5} 
\multicolumn{1}{l}{} & Day                & Night               & DUC1               & DUC2               \\ \midrule
MASt3R top1          & 77.5 / 89.0 / 97.9 & 58.2 / 72.4 / 87.8  & 43.9 / 60.1 / 68.7 & 38.2 / 54.2 / 55.0 \\
MASt3R top20         & 88.2 / 95.1 / 99.6 & 72.4 / 93.9 / 99.0  & 63.6 / 82.8 / 90.9 & 69.5 / 89.3 / 89.3 \\
MASt3R top40         & 89.2 / 95.4 / 99.8 & 75.5 / 91.8 / 100.0 & 63.6 / 84.3 / 93.4 & 72.5 / 91.6 / 92.4 \\
Speedy top1          & 77.2 / 89.2 / 98.1 & 57.1 / 73.5 / 87.8  & 43.9 / 60.1 / 68.7 & 38.9 / 54.2 / 55.0 \\
Speedy top20         & 88.2 / 95.4 / 99.6 & 71.4 / 92.9 / 98.0  & 63.6 / 82.3 / 91.4 & 69.5 / 89.3 / 89.3 \\
Speedy top40         & 89.1 / 95.5 / 99.8 & 74.5 / 90.8 /100.0  & 63.1 / 84.3 / 93.4 & 72.5 / 91.6 / 92.4 \\ \bottomrule
\end{tabular}%
}
\resizebox{\textwidth}{!}{%
\begin{tabular}{@{}cccccccc@{}}
\toprule
\multicolumn{1}{l}{\multirow{2}{*}{}} & \multicolumn{7}{c}{7-Scenes~\cite{shotton2013scene}}                                                   \\ \cmidrule(l){2-8} 
\multicolumn{1}{l}{} & Chess       & Fire        & Heads       & Office      & Pumpkin     & Kitchen     & Stairs      \\ \midrule
MASt3R top1          & 2.33 / 0.79 & 2.16 / 0.80 & 1.20 / 0.82 & 3.10 / 0.92 & 4.10 / 1.07 & 3.71 / 1.28 & 3.46 / 0.95 \\
Ours top1            & 2.33 / 0.79 & 2.16 / 0.80 & 1.20 / 0.82 & 3.10 / 0.92 & 4.11 / 1.07 & 3.71 / 1.28 & 3.46 / 0.95 \\ \bottomrule
\end{tabular}%
}
\end{table*}

\begin{table*}[]
\centering
\caption{Computational efficiency test on Aachen Day-Night~\cite{Zhang2021Reference} and InLoc~\cite{taira2018inloc} (left) and on the 7-Scenes~\cite{shotton2013scene} (right). Lower numbers are better.}
\label{tab:3}
\begin{minipage}{.5\textwidth}
\centering
\resizebox{\textwidth}{!}{%
\begin{tabular}{@{}lcccccc@{}}
\toprule
\multicolumn{1}{c}{}             & \multicolumn{1}{l}{} & Encoder\(\downarrow\)   & Decoder\(\downarrow\) & Head\(\downarrow\) & FastNN\(\downarrow\)             \\ \midrule
\multicolumn{1}{c}{Aachen Day}   & MASt3R top1          & 57.21    & 39.71  & 22.97            & 78.27       \\
 & MASt3R top20 & 969.05  & 571.39  & 388.90 & 1140.17   \\
 & MASt3R top40 & 2206.40 & 1586.96 & 871.41 & 3033.58   \\
 & Ours top1    &  \textbf{30.24}   &  \textbf{23.28}   &  \textbf{15.11}  &  \textbf{22.87}      \\
 & Ours top20   & 599.09  & 475.06  & 283.99 & 441.07     \\
 & Ours top40   & 1204.44 & 928.27  & 570.28 & 895.47     \\ \midrule
\multicolumn{1}{c}{Aachen Night} & MASt3R top1          & 85.80    & 63.17  & 30.19            & 130.98     \\
 & MASt3R top20 & 1037.35 & 680.83  & 414.63 & 1316.06    \\
 & MASt3R top40 & 2198.81 & 1519.02 & 866.82 & 2996.83    \\
 & Ours top1    &  \textbf{31.57}   &  \textbf{23.53}   &  \textbf{16.40}  &  \textbf{22.64}      \\
 & Ours top20   & 617.19  & 467.74  & 299.73 & 459.95     \\
 & Ours top40   & 1195.93 & 935.39  & 574.23 & 886.05     \\ \midrule
\multicolumn{1}{c}{InLoc}        & MASt3R top1          & 65.89    & 32.19  & 28.69            & 1766.62   \\
 & MASt3R top20 & 1144.30 & 812.05  & 451.10 & 39260.08    \\
 & MASt3R top40 & 2101.70 & 1266.11 & 829.22 & 70941.91    \\
 & Ours top1    &  \textbf{30.63}   &  \textbf{23.50}   &  \textbf{15.77}  &  \textbf{770.05}      \\
 & Ours top20   & 591.42  & 457.33  & 307.74 & 15909.04    \\
 & Ours top40   & 1197.402 & 897.26 & 616.06 & 32101.84   \\ \bottomrule
\end{tabular}%
}
\end{minipage}
\hfill
\begin{minipage}{.49\textwidth}
\centering
\resizebox{\textwidth}{!}{%
\begin{tabular}{@{}ccccccc@{}}
\toprule
7-Scenes              & \multicolumn{1}{l}{} & Encoder\(\downarrow\) & Decoder\(\downarrow\) & Head\(\downarrow\)    & FastNN\(\downarrow\)       \\ \midrule
Chess                & MASt3R top1          & 53.00   & 31.70   & 21.35   & 1702.78   \\
\multicolumn{1}{l}{} & Ours top1            & \textbf{28.66}   &  \textbf{22.11}   &  \textbf{15.23}   &  \textbf{761.10}    \\
Fire                 & MASt3R top1          & 52.76   & 31.91   & 21.00   & 1811.34   \\
\multicolumn{1}{l}{} & Ours top1            & \textbf{28.57}   &  \textbf{21.89}   &  \textbf{15.16}   &  \textbf{797.47}    \\
Heads                & MASt3R top1          & 53.70   & 33.13   & 21.67   & 1708.31   \\
\multicolumn{1}{l}{} & Ours top1            & \textbf{29.73}   &  \textbf{22.29}   &  \textbf{15.75}   &  \textbf{748.81}    \\
Office               & MASt3R top1          & 53.40   & 31.25   & 20.98   & 1704.45   \\
\multicolumn{1}{l}{} & Ours top1            & \textbf{28.23}   &  \textbf{21.71}   &  \textbf{15.03}   &  \textbf{775.07}    \\
Pumpkin              & MASt3R top1          & 53.18   & 32.18   & 21.89   & 1749.20   \\
\multicolumn{1}{l}{} & Ours top1            & \textbf{29.47}   &  \textbf{22.11}   &  \textbf{15.33}   &  \textbf{770.75}    \\
Kitchen              & MASt3R top1          & 54.32   & 36.00   & 22.03   & 1934.92   \\
\multicolumn{1}{l}{} & Ours top1            & \textbf{28.07}   &  \textbf{21.54}   &  \textbf{15.03}   &  \textbf{792.70}    \\
Stairs               & MASt3R top1          & 52.75   & 30.85   & 20.63   & 1696.28   \\
\multicolumn{1}{l}{} & Ours top1            & \textbf{31.39}   &  \textbf{23.52}   &  \textbf{16.71}   &  \textbf{775.33}    \\ \bottomrule
\end{tabular}%
}
\end{minipage}
\end{table*}

\begin{table*}[]
\centering
\caption{Relative pose estimation accuracy remains stable while optimization techniques are applied incrementally.}
\label{tab:4}
\resizebox{\textwidth}{!}{%
\begin{tabular}{@{}ccccccccc@{}}
\toprule
\multicolumn{1}{l}{\multirow{2}{*}{}} & \multicolumn{4}{c}{ScanNet1500~\cite{dai2017scannet}} & \multicolumn{4}{c}{MegaDepth1500~\cite{li2018megadepth}} \\ \cmidrule(l){2-9} 
\multicolumn{1}{l}{}  & AUC@5 & AUC@10 & AUC@20 & mAA   & AUC@5 & AUC@10 & AUC@20 & mAA   \\ \midrule
MASt3R                & 34.51 & 57.31  & 74.50  & 55.44 & 39.87 & 54.93  & 66.88  & 53.89 \\
+ FlashMatch          & 34.41 & 57.24  & 74.37  & 55.34 & 38.93 & 54.46  & 66.93  & 53.44 \\
+ GraphFusion            & 34.24 & 57.32  & 74.52  & 55.36 & 39.30 & 55.00  & 66.98  & 53.76 \\
+ FastNN-Light        & 34.32 & 57.15  & 74.30  & 55.25 & 39.27 & 54.56  & 66.52  & 53.45 \\
+ HybridCast (Speedy MASt3R) & 34.18 & 57.17  & 74.37  & 55.24 & 38.83 & 54.74  & 66.88  & 53.48 \\ \bottomrule
\end{tabular}%
}
\end{table*}

\begin{table}[]
\centering
\caption{Computational efficiency increases when optimization techniques are applied incrementally. Lower numbers are better}
\label{tab:5}
\resizebox{\columnwidth}{!}{%
\begin{tabular}{@{}lcccccc@{}}
\toprule
                              & \multicolumn{1}{l}{} & Encoder\(\downarrow\) & Decoder\(\downarrow\) & Head\(\downarrow\)   & FastNN\(\downarrow\)      \\ \midrule
\multicolumn{1}{c}{ScanNet1500~\cite{dai2017scannet}}   & MASt3R               & 52.79   & 31.01   & 20.72  & 115.69   \\
                              & + FlashMatch         & 27.64   & 21.43   & 20.65  & 125.17   \\
                              & + GraphFusion           & 27.99   & 21.42   & 15.13  & 116.54   \\
                              & + FastNN-Light       & 28.41   & 21.62   & 15.59  & 70.19    \\
                              & + HybridCast (Speedy MASt3R)                &  \textbf{27.50}   &  \textbf{21.43}   &  \textbf{15.04}  &  \textbf{42.64}    \\ \midrule
\multicolumn{1}{c}{MegaDepth1500~\cite{li2018megadepth}} & MASt3R               & 52.34   & 30.85   & 20.77  & 114.17   \\
                              & + FlashMatch         & 27.75   & 21.37   & 20.74  & 114.02   \\
                              & + GraphFusion           & 27.74   & 21.41   & 15.18  & 121.80   \\
                              & + FastNN-Light       & 27.49   & 21.37   & 15.09  & 70.66    \\
                              & + HybridCast (Speedy MASt3R)                &  \textbf{27.76}   &  \textbf{21.38}   &  \textbf{15.15}  &  \textbf{43.60}    \\ \bottomrule
\end{tabular}%
}
\end{table}


\subsection{Visual Localization}
\label{sec:visual_localization}
In this scenario, we evaluate the accuracy of estimated absolute pose across three datasets: Aachen Day-Night~\cite{Zhang2021Reference}, InLoc~\cite{taira2018inloc}, and 7-Scenes~\cite{shotton2013scene}. The Aachen dataset consists of 824 daytime and 98 nighttime query images, along with 5,235 reference images captured in the historic city center of Aachen, Germany. The InLoc dataset presents challenges in estimating the correct pose for 356 hand-captured query images, given a database of 4,681 RGB images with significant visual differences. The 7-Scenes dataset includes seven distinct indoor environments, each containing a varying number (1,000–5,000) of query images.  

We evaluate localization performance by measuring the percentage of successfully localized images within three thresholds: (0.25m/2\textdegree), (0.5m/5\textdegree), and (5m/10\textdegree) for Aachen, and (0.25m/10\textdegree), (0.5m/10\textdegree), and (1m/10\textdegree) for InLoc. For 7-Scenes, we report the median translation and rotation errors in meters and degrees, respectively. Additionally, we assess computational efficiency across all three datasets, analyzing the processing time of each module (in ms) required for localizing a single query image.

For each query image, we evaluate localization performance using the top 1, top 20, and top 40 retrieved images. As shown in \Cref{tab:2}, Speedy MASt3R improves localization accuracy as more retrieved images are provided, similar to MASt3R. \Cref{tab:3} further demonstrates the enhanced computational efficiency of Speedy MASt3R across all three datasets. Notably, Speedy MASt3R achieves greater time savings with an increasing number of retrieved images. For example, in the case of InLoc (top 40), Speedy MASt3R reduces the running time of each MASt3R module by 0.904s, 0.368s, 0.213s, and 38.840s, resulting in a total time savings of 40.326s.

\subsection{Ablation study}
To assess the impact of each optimization technique on the modules, we incrementally apply FlashMatch, GraphFusion, FastNN-Light, and HybridCast one by one. We evaluate relative pose estimation quality using AUC@5/10/20 and mAA (\Cref{tab:4}) and measure the running time of each module in ms (\Cref{tab:5}) on the ScanNet1500~\cite{dai2017scannet} and MegaDepth1500~\cite{li2018megadepth} benchmarks.  

We observe that on both ScanNet1500 and MegaDepth1500 benchmark, Speedy MASt3R maintains the same accuracy with vanilla MASt3R; the minor differences are statistically insignificant.
In terms of computational efficiency, each technique effectively reduces the running time of the targeted modules. On the ScanNet1500 benchmark, FlashMatch, GraphFusion, FastNN-Light, and HybridCast reduce processing time by 25.15ms/9.58ms in the Encoder/Decoder, 5.59ms in the Head, 45.5ms in FastNN, and an additional 27.55ms in FastNN, respectively. Similarly, on the MegaDepth1500 benchmark, these techniques reduce processing time by 24.59ms/9.48ms in the Encoder/Decoder, 5.59ms in the Head, 43.51ms in FastNN, and an additional 27.06ms in FastNN.

\section{Conclusion}
\label{sec:conclusion}

In this work, we introduced \textit{Speedy MASt3R}, a post-training optimization framework designed to accelerate the inference speed of the MASt3R image matching model while maintaining its state-of-the-art accuracy. Speedy MASt3R integrates multiple optimizations, including \textit{FlashMatch}, \textit{GraphFusion}, \textit{FastNN-Lite}, and \textit{HybridCast}, each targeting key computational bottlenecks in the original MASt3R pipeline. These enhancements enable a significant reduction in inference time~(from 198 ms to 91 ms per image pair) without compromising matching performance.

Through extensive evaluations on benchmark datasets such as ScanNet1500, MegaDepth1500, Aachen Day-Night, InLoc, and 7-Scenes, we demonstrate that Speedy MASt3R preserves the theoretical guarantees as well as practical performance of fast reciprocal matching with significant improvement in inference time (more than $54$ percentage).


Our findings underscore the critical need to enhance MASt3R’s efficiency, given its growing adoption as a state-of-the-art image matching model in 3D vision. While MASt3R delivers exceptional accuracy even in challenging scenarios, its computational overhead presents a significant challenge. Speedy MASt3R tries to address this crucial limitation and represents a significant step in that direction by significantly accelerating MASt3R’s inference (more than 50\%) without compromising its robust performance.


\ifreview
\else
    \section*{Acknowledgments}
    Supported by the Intelligence Advanced Research Projects Activity (IARPA) via Department of Interior/ Interior Business Center (DOI/IBC) contract number 140D0423C0076. The U.S. Government is authorized to reproduce and distribute reprints for Governmental purposes notwithstanding any copyright annotation thereon. Disclaimer: The views and conclusions contained herein are those of the authors and should not be interpreted as necessarily representing the official policies or endorsements, either expressed or implied, of IARPA, DOI/IBC, or the U.S. Government.
\fi

{
    \small
    \bibliographystyle{ieeenat_fullname}  %
    \bibliography{references}

}
\end{document}

%% file: preamble.tex
\usepackage{graphicx}
\usepackage{times}
\usepackage{epsfig}
\usepackage{multirow}
\usepackage{booktabs}

